\documentclass[letterpaper, 10 pt, conference]{ieeeconf}  % Comment this line out if you need a4paper

\IEEEoverridecommandlockouts                              % This command is only needed if 
\usepackage{amsmath,amssymb,amsfonts}
\usepackage{algorithmic}
\usepackage{graphicx}
\usepackage{epstopdf}
\usepackage{epsfig,adjustbox}
\usepackage{stfloats}
\usepackage{stackengine}
\usepackage{textcomp}
\usepackage{xcolor}
\usepackage{caption}
\usepackage{subcaption}
\usepackage{biblatex} %Imports biblatex package
\title{\LARGE \bf
Learning Vision-based Robotic Manipulation Tasks Sequentially in Offline Reinforcement Learning Settings
}

\author{Sudhir Pratap Yadav$^{1}$, Rajendra Nagar$^{2}$, and Suril V. Shah$^{3}$% <-this % stops a space
\thanks{*This work was done with collaboration of IIT Jodhpur and iHub-Drishti Foundation, IIT Jodhpur }% <-this % stops a space
\thanks{$^{1}$Sudhir Pratap Yadav, iHub Drishti Foundation
        {\tt\small sudhiry@ihub-drishti.ai}}%
\thanks{$^{2}$Rajendra Nagar, IIT Jodhpur
        {\tt\small rn@iitj.ac.in}}%
\thanks{$^{3}$Suril V. Shah, IIT Jodhpur
        {\tt\small surilshah@iitj.ac.in}}%
}

\begin{document}

\maketitle
\thispagestyle{empty}
\pagestyle{empty}

%%%%%%%%%%%%%%%%%%%%%%%%%%%%%%%%%%%%%%%%%%%%%%%%%%%%%%%%%%%%%%%%%%%%%%%%%%%%%%%%
\begin{abstract}
With the rise of deep reinforcement learning (RL) methods, many complex robotic manipulation tasks are being solved. However, harnessing the full power of deep learning requires large datasets. Online-RL does not suit itself readily into this paradigm due to costly and time-taking agent environment interaction. Therefore recently, many offline-RL algorithms have been proposed to learn robotic tasks. But mainly, all such methods focus on a single task or multi-task learning, which requires retraining every time we need to learn a new task. Continuously learning tasks without forgetting previous knowledge combined with the power of offline deep-RL would allow us to scale the number of tasks by keep adding them one-after-another. In this paper, we investigate the effectiveness of regularisation-based methods like synaptic intelligence for sequentially learning image-based robotic manipulation tasks in an offline-RL setup. We evaluate the performance of this combined framework against common challenges of sequential learning: catastrophic forgetting and forward knowledge transfer. We performed experiments with different task combinations to analyze the effect of task ordering. We also investigated the effect of the number of object configurations and density of robot trajectories. We found that learning tasks sequentially helps in the propagation of knowledge from previous tasks, thereby reducing the time required to learn a new task. Regularisation based approaches for continuous learning like the synaptic intelligence method although helps in mitigating catastrophic forgetting but has shown only limited transfer of knowledge from previous tasks.

\end{abstract}

%%%%%%%%%%%%%%%%%%%%%%%%%%%%%%%%%%%%%%%%%%%%%%%%%%%%%%%%%%%%%%%%%%%%%%%%%%%%%%%%
\section{INTRODUCTION}
Robots now have the capability to learn many single manipulation tasks using deep Reinforcement Learning (RL), such as pick-place \cite{berscheid2020self}, peg-in-hole \cite{yasutomi2021peg}, Cloth folding \cite{lee2020learning}, and tying rope knots \cite{nair2017combining}. Multitask RL has also been applied
successfully to learn robotic manipulation tasks \cite{gupta2021reset}, \cite{kalashnikov2021scaling}. Number of tasks and task-data distribution are kept fixed in the case of multi-task RL. Therefore, agent has to be trained from scratch whenever it needs to learn a new task, even if there is a substantial overlap between tasks. Scaling this approach to learn all manipulation tasks at par with humans is not feasible. Humans use the experience of previous tasks for learning a new task and do not need to learn from the start. The sequential (or continual) learning approach tries to address this problem by providing a framework where an agent can learn new tasks one-after-another without starting from scratch. We use offline-RL as the base framework to learn a single image-based robotic-manipulation task and then use a regularisation based continual learning approach for learning tasks sequentially. This combined framework forms main contribution of this work.

\begin{figure}[t]
  \centering
  \includegraphics[width=1\linewidth]{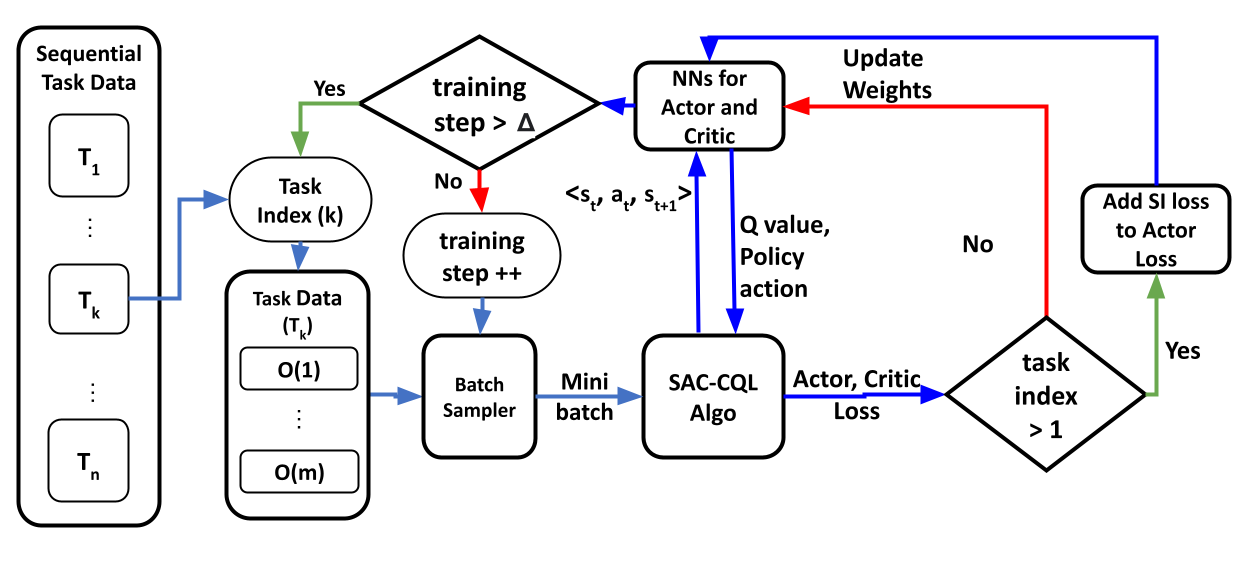}
  \caption{Block Diagram of SAC-CQL-SI method for Sequential Learning}
  \label{fig:sac_cql_si}
\end{figure}

\subsection{Related Work}
Most work in sequential task learning is focused on classification-based tasks using typical classification datasets such as MNIST, CIFAR and their variations \cite{goodfellow2013empirical}, \cite{mallya2018packnet}. Some works in continual reinforcement Learning setup use Atari-games \cite{kirkpatrick2017ewc}. Other try to extend continual RL to GYM environments \cite{DBLP:journals/corr/BrockmanCPSSTZ16}. Recent work in \cite{singh2020cog} uses Offline-RL for solving manipulation tasks using image observations alone. While this work focuses on generalizing to novel initial conditions but does not attempt sequential task learning. On the other hand, our work is about learning tasks sequentially with only the current task data available for learning.
Some very recent works try to apply continual RL on robotics manipulation tasks \cite{wolczyk2021continual}, \cite{caccia2022task}. \cite{wolczyk2021continual} introduces a continual learning benchmark for robotic manipulation tasks. It gives baselines for major continual learning methods over these robotics tasks in online RL settings using soft actor-critic (SAC) method \cite{haarnoja2018soft}. But this work focuses on online-continual RL with low-dimensional observation space such as joint and task space data as they assume full access to the simulator. While our work focuses on Offline RL with high-dimensional observation space (image) in sequential learning of robotics manipulation tasks.

In the sequential learning setup based on deep RL, neural-networks (NN) are prone to change in data-distribution. Hence, its accuracy on previous tasks drops significantly when it is trained on a new task. This problem is actively studied under the name of catastrophic forgetting. More broadly, in every connectionist model of memory and computation problem of \textbf{stability-plasticity} exists. This means the network needs to be flexible enough to accommodate new information and simultaneously not forget previous information, as discussed in \cite{french1999catastrophic}.
Many solutions have been suggested to mitigate this problem. We place these under two major categories architectural and penalty based. In the architectural type of solutions, relevant changes are made in the neural network architecture without changing the loss function. For example, Progressive neural network \cite{rusu2016progressivenn} uses multiple parrallel paths with lateral connenctions, and policy distillation \cite{rusu2015policydistil} distills the policy learned by larger network into smaller without loss of performance. On the other hand, penalty based methods put penalties on neural network parameters so that they stay close to solution of previous task. Two important work in this regards are Elastic Weight Consolidation (EWC) \cite{kirkpatrick2017ewc} and Synaptic Intelligence \cite{zenke2017synaptic}. EWC gives regularisation based solution for catastrophic forgetting, but computation for the importance of parameters is not local. In this paper we use the approach proposed in Synaptic Intelligence \cite{zenke2017synaptic} because of its local measure of importance for synapses (weights of Neural Network) as the nature of local computation helps in keeping the solution independent of the particularities of the problem hence making it more general.

To the best of our knowledge, this is the first work investigating sequential learning for image based robotic task manipulation in offline RL settings. In this paper, we focus on two sequential learning challenges: catastrophic forgetting and forward knowledge transfer. We analyse the effect of task-ordering and number of object configurations on forgetting and forward knowledge transfer between tasks.

\section{LEARNING IMAGE BASED ROBOTIC MANIPULATION TASKS SEQUENTIALLY}
In this section, we formulate our RL agent and environment interaction setup to learn robotic manipulation tasks. We then discuss the problem of sequential task learning and present an approach to solve this problem.

\subsection{RL formulation for Learning Image Based Robotic Manipulation Tasks}
Agent and environment interaction is formally defined by the Markov Decision Process (MDP) concept. A Markov Decision Process is a discrete-time stochastic control process. In RL,  we formally define the MDP as a tuple $\langle\mathcal{S},\mathcal{A}, \mathsf{P},r, \gamma\rangle$. Here,  $\mathcal{S}$ is a finite set of states, $\mathcal{A}$ is a finite set of actions, $\mathsf{P}$ is the state transition probability matrix, $r$ is the reward for a given state-action pair and $\gamma$ is the discount factor.
A stochastic policy is defined as a distribution over actions given the states, i.e., the probability of taking each action for every state. $\pi(\mathbf{a}|\mathbf{s}) = \mathbb{P}[\mathcal{A}_t=\mathbf{a}|\mathcal{S}_t=\mathbf{s}]$.

\textbf{RL formulation:} We formulate the vision-based robotic manipulation tasks using the deep RL framework as below.  
\begin{itemize}
\item \textbf{Environment:} It consists of WidowX 250 five-axes robot arm equipped with a gripper. We place a table in front of the robot. Every task consists of an object placed on the table, which needs to be manipulated to complete the task successfully. We place a camera in the environment in eye-to-hand configuration.
\item \textbf{State:} The state $\mathbf{s}_t$ represents the RGB image of the environment captured at time step $t$. We use capture images of size $48\times48\times3$.
\item \textbf{Action:} We define the action at the time step $t$ as a 7 dimensional vector $\mathbf{a}_t=\begin{bmatrix}\Delta X_t&\Delta O_t&g_t\end{bmatrix}^\top$. Here, $\Delta X_t\in\mathbb{R}^3$, $\Delta O_t\in\mathbb{R}^3$, $g_t\in\{0,1\}$ denotes the change in position, change in orientation, and gripper command (open/close), respectively at time step $t$.  
\item \textbf{Reward:} The reward $r(\mathbf{s}_t,\mathbf{a}_t)\in\{0,1\}$ is a binary variable which is equal to 1 if the task is successful and 0, otherwise. Reward is given at each time step. 
\end{itemize}
The reward is kept simple and not shaped according to the tasks so that the same reward framework can be used while scaling for large number of tasks. Also, giving reward at each time step, instead at the end of the episode, makes the sum of rewards during an episode dependent on time steps. Therefore, if the agent completes a task in fewer steps, its total reward for that episode will be more.

\subsection{Sequential Learning Problem and Solution}
We define the sequential tasks learning problem as follows. The agent is required to learn $N$ number of tasks but with the condition that tasks will be given sequentially to the agent and not simultaneously. Therefore, when the agent is learning to perform a particular task, it can only access the data of the current task. This learning process reassembles how a human learns. Let a sequence of robotic manipulation tasks $T_1, T_2, ..., T_N$ is given. We assume that each task has the same type of state and action space. Each task has its own data in typical offline reinforcement learning format $\langle\mathbf{s}_t, \mathbf{a}_t, \mathbf{r}_t, \mathbf{s}_{t+1}\rangle$. The agent has to learn a policy $\pi$, a mapping from state to action, for every task. If we naively train a neural network in this fashion problem of catastrophic forgetting will occur, which means performance on the previous task will decrease drastically as soon as the neural network starts learning a new task.

We use a regularisation based approach presented in \cite{zenke2017synaptic} to mitigate the problem of catastrophic forgetting. Figure \ref{fig:sac_cql_si} provides the overall framework we use to solve this problem. Each task data is given one by one to the algorithm, which then starts training for the current task. First, a mini-batch is sampled from this current-task data and passed to the SAC-CQL algorithm (described in the next section), which then calculates actor (Q-loss) and critic (policy) loss. If the task-index is greater than one then we add a quadratic regularisation as defined in \cite{zenke2017synaptic} to the actor loss to reduce forgetting. Then, these losses are used to update neural networks, which represents policy (actor network) and Q-value function (critic network). After the current task is successfully learned next task data comes, and this process is repeated until all tasks are learned.

\section{INTEGRATING SEQUENTIAL TASK LEARNING WITH OFFLINE RL}
In this section, we discuss the SAC-CQL \cite{kumar2020conservative} offline algorithm and its implementation details. We then discuss the SI regularisation method for continual learning and provide details to integrate these methods to learn sequential tasks.

\subsection{SAC-CQL algorithm for Offline RL}
There are two frameworks, namely online and offline learning, to train an RL agent. In the case of an online-RL training framework, an RL agent interacts with the environment to collect experience, update itself (train), interact again, and so on. In simple terms, the environment is always available for the RL agent to evaluate itself and improve further. This interaction loop is repeated for many episodes during training until the RL agent gets good enough to perform the task successfully. While in offline RL settings, we collect data once and are no more required to interact with the environment. This data can be collected by executing a hand-designed policy or can be obtained by a human controlling the robot (human-demonstration). Data is a sequence of $\langle\mathbf{s}_t, \mathbf{a}_t, \mathbf{r}_t, \mathbf{s}_{t+1}\rangle$ tuples.

In recent years the SAC (soft-actor critic) \cite{haarnoja2018soft} has emerged as the most robust algorithm for training RL agents in continuous action space (when action is a real vector), which typically is a case in robotics.
SAC is an off-policy entropy based actor-critic method for continuous action MDPs. Entropy based methods add an additional entropy term to the existing optimisation goal of maximising expected reward. In addition to maximising expected reward, the RL agent also needs to maximise the entropy of the overall policy. This helps in making the policy inherently exploratory and not stuck inside a local minima. Haarnoja \emph{et al.} \cite{haarnoja2018soft} define the RL objective in maximum entropy RL settings as in \eqref{eq:maximum_entropy_objective}.

\begin{equation} \label{eq:maximum_entropy_objective}
J(\pi) = \sum \limits_{t=0}^{T} \mathbb{E}_{(\mathbf{s}_t, \mathbf{a}_t)\sim\rho_\pi}[r(\mathbf{s}_t, \mathbf{a}_t)+\alpha \mathcal{H}(\pi(\cdot|\mathbf{s}_t))].
\end{equation} 
Here, $\rho_\pi(\mathbf{s}_t, \mathbf{a}_t)$ denotes the joint distribution of the state and actions over all trajectories of the agent could take and $\mathcal{H}(\pi(\cdot|\mathbf{s}_t))$ is the entropy of the policy for state $\mathbf{s}_t$ as defined in \eqref{eq:entropy_of_policy}.
\begin{equation} \label{eq:entropy_of_policy}
\mathcal{H}(\pi(\cdot|\mathbf{s}_t)) = \mathbb{E}[-\text{log}(f_\pi(\cdot|\mathbf{s}_t))].
\end{equation}
Here,  $\pi(\cdot|\mathbf{s}_t)$ is a probability distribution over actions and $f_\pi(\cdot|\mathbf{s}_t)$ is the density function of the policy $\pi$. $\alpha$ is the temperature parameter controlling the entropy in the policy.

SAC provides an actor-critic framework where policy is separately represented by the actor and critic only helps in improving the actor, thus limiting its role only to training. We use CNNs to represent both actor and critic, and instead of using a single Q-value network for the critic, we use two Q-value networks and take their minimum to better estimate Q-value as proposed in \cite{van2016deep}. To stabilize the learning, we use two more neural-network to represent target Q-values for each critic network, as described in DQN \cite{mnih2013playing}. $\phi$, $\theta_1$, $\theta_2$, $\hat{\theta}_1$ and $\hat{\theta}_2$ represents parameters of policy network, 2 Q-value networks and 2 target Q-value networks for critic respectively. Therefore in total, we use 5 CNNs to implement the SAC algorithm.

Our CNN architecture is similar to \cite{singh2020cog} except for the multi-head part, which is a single layer neural-network for each head. Q-value network takes state and action as input and directly gives Q-value. We use \emph{tanh-guassian} policy, as used in \cite{singh2020cog}. Since we use stochastic policy thus, the policy network takes the state as input and outputs the mean and standard deviation of the gaussian distribution of each action. Action is then sampled from this distribution and passed through \emph{tanh} function to bound actions between $(-1, 1)$. Equation \eqref{eq:Q_target} defines the target Q-value which is then used in \eqref{eq:Q_loss} to calculate Q-loss for each critic networks. Equation \eqref{eq:P_loss} defines policy-loss for actor network. These losses are then used to update actor and critic networks using adam \cite{kingma2014adam} optimisation algorithm. 
\begin{multline} \label{eq:Q_target}
\hat{Q}_{\bar{\theta}_1,\bar{\theta}_2}(\mathbf{s}_{t+1}, \mathbf{a}_{t+1}) =  \mathbf{r}_t \\
+\gamma \mathbb{E}_{(\mathbf{s}_{t+1} \sim \mathcal{D}, \mathbf{a}_{t+1} \sim \pi_\phi(\cdot|\mathbf{s}_{t+1}))}[ \\
\text{ min}[Q_{\bar{\theta}_1}(\mathbf{s}_{t+1}, \mathbf{a}_{t+1}), Q_{\bar{\theta}_2}(\mathbf{s}_{t+1}, \mathbf{a}_{t+1})] \\
-\alpha \text{log}(\pi_\phi(\mathbf{a}_{t+1}|\mathbf{s}_{t+1}))]      
\end{multline}
\begin{align} \label{eq:Q_loss}
J_Q({\theta_i}) &= \frac{1}{2} \mathbb{E}_{(\mathbf{s}_t, \mathbf{a}_t) \sim \mathcal{D}}[(\hat{Q}_{\bar{\theta}_i,\bar{\theta}_2}(\mathbf{s}_{t+1}, \mathbf{a}_{t+1}) - Q_{\theta_1}(\mathbf{s}_t, \mathbf{a}_t))^2].
\end{align}
\begin{multline} \label{eq:P_loss}
J_\pi(\phi) =\mathbb{E}_{(\mathbf{s}_t \sim \mathcal{D}, \mathbf{a}_t \sim \pi_\phi(\cdot|\mathbf{s}_t))}[\alpha \text{log}(\pi_\phi(\mathbf{a}_t|\mathbf{s}_t)) \\
-\text{ min}[Q_{\theta_1}(\mathbf{s}_t, \mathbf{a}_t^\pi), Q_{\theta_2}(\mathbf{s}_t, \mathbf{a}_t^\pi)]]
\end{multline}
Here, $i\in\{1,2\}$, $\mathbf{a}_t^\pi$ is the action sampled from policy $\pi_\phi$ for state $\mathbf{s}_t$ and $\mathcal{D}$ represents the current task data. For offline-RL, we use the non-Lagrange version of the conservative Q-learning (CQL) approach proposed in \cite{kumar2020conservative} as it only requires adding a regularisation loss to already well-established continuous RL methods like Soft-Actor Critic. This loss function is defined in \eqref{eq:cql_loss}.
\begin{multline} \label{eq:cql_loss}
J_Q^{\text{total}}({\theta_i}) = J_Q({\theta_i}) \\
+\alpha_{\text{cql}} \mathbb{E}_{\mathbf{s}_t \sim \mathcal{D}}[\text{log}\sum\limits_{\mathbf{a}_t} \text{exp}(Q_{\theta_i}(\mathbf{s}_t, \mathbf{a}_t))-\mathbb{E}_{\mathbf{a}_t \sim \mathcal{D}}[Q_{\theta_i}(\mathbf{s}_t, \mathbf{a}_t)]]
\end{multline}
Here, $i\in\{1,2\}$, $\alpha_{\text{cql}}$ control the amount of CQL-loss to be added to Q-loss to penalize actions that are too far away from the existing trajectories, thus keeping the policy conservative in the sense of exploration.
\subsection{Applying Synaptic Intelligence in Offline RL}
Synaptic intelligence is a regularisation based algorithm proposed in \cite{zenke2017synaptic} for sequential task learning. It regularises the loss function of a task with a quadratic loss function as defined in \eqref{eq:si_loss} to reduce catastrophic forgetting. 
\begin{equation} \label{eq:si_loss}
 L_\mu = \sum \limits_{k} \Omega^\mu_k(\tilde{\phi}_k-\phi_k)^2
\end{equation}
Here, $L_\mu$ is the SI loss for the current task being learned with index $\mu$, $\phi_k$ is $k$-th weight of the policy network, and $\tilde{\phi}_k$ is the reference weight corresponding to policy network parameters at the end of the previous task. $\Omega^\mu_k$ is per-parameter regularisation strength for more details on how to calculate $\Omega^\mu_k$ refer to \cite{zenke2017synaptic}.
SI algorithm penalizes neural network weights based on their contributions to the change in the overall loss function. Weights that contributed more to the previous tasks are penalized more and thus do not deviate much from their original values, while other weights help learn new tasks. SI defines importance of weights as the sum of the gradients over the training trajectory, as this approximates contribution to the reduction in the overall loss function.
We use a similar approach to apply SI to Offline-RL as presented in \cite{wolczyk2021continual}. Although the authors didn't use SI or offline-RL, the approach is similar to applying any regularisation based continual learning method for the actor-critic RL framework.
We regularise the actor to reduce forgetting on previous tasks while learning new tasks using offline reinforcement learning. We add quadratic loss as defined in \cite{zenke2017synaptic} to the policy-loss term in the SAC-CQL algorithm. So now over-all policy-loss becomes as described in \eqref{eq:actor_regularisation}
\begin{equation} \label{eq:actor_regularisation}
J_{\pi}^{\text{total}}(\phi) = J_\pi(\phi) + c L_\mu
\end{equation}
Here, $c$ is regularisation strength. Another aspect of continual learning is finding a way to provide the current task index to the neural network. There are many approaches to tackle this problem, from 1-hot encoding to recognizing the task from context. We chose the most straightforward option of a multi-head neural network. Each head of the neural network represents a separate task. Therefore we simply select the head for a given task. For training each task we keep a fixed compute budget of 100k of gradient-steps. 

\section{EXPERIMENTS, RESULTS AND DISCUSSION}
In this section we first discuss the RL environment setup and provide details of data collection for offline RL. Further, we evaluate performance of SI with varying number of object configurations and densities for different task ordering.

\subsection{Experimental Setup}
Our experimental setup is based on a simulated environment, Roboverse, used in \cite{singh2020cog}. It is a GYM \cite{brockman2016openai} like environment based upon open-source physics simulator py-bullet \cite{erwin2016pybullet}. We collected data for three tasks using this simulated environment.

\textbf{Object Space and Tasks:} We define object space as a subset of the workspace of the robot where the target object of the task is to be placed. In our case, it is a rectangular area on the table in front of the robot. The target object is randomly placed in the object-space when initializing the task. We selected the following 3 tasks for all our experiments with some similarities.

1) \emph{Press Button:} Button is placed in the object-space. The objective of the task is to press the button. This is easiest task as the robot only needs to learn to reach the object.

\begin{figure}
\centering
  \stackunder{\includegraphics[width=0.24\linewidth]{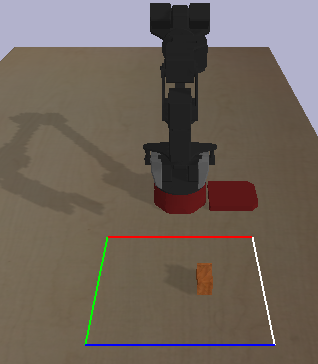}}{(a)}
  \stackunder{\includegraphics[width=0.74\linewidth]{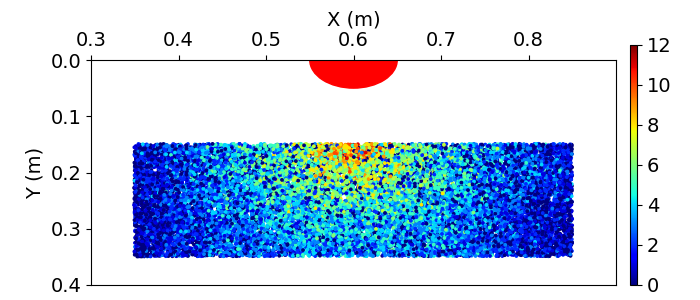}}{(b)}
\caption{Object space and reward distribution for pick-shed task with area of size of $1000\text{ cm}^2$  and density of 20 object configurations per $\text{ cm}^2$. (a) Object Space. (b) Reward distribution (cumulative reward along sample trajectories) of pick-shed task (area=1000, density=20). Red semicircle on top represents the robot location }
 \label{fig:obj_spcae_reward_dist}
\end{figure}

2) \emph{Pick Shed:} The objective of this task is to pick the object successfully. Thus, robot also needs to learn to close the gripper apart from reaching the object. Figure \ref{fig:obj_spcae_reward_dist}(a) shows the object space of task pick-shed.

3) \emph{Open Drawer:} The objective of this task is to open the drawer.

\textbf{Data Collection:} For each task we collect 6 datasets by varying the area ($\text{40, 360 and 1000 cm}^2$) and density ($\text{10 and 20 object configurations per cm}^2$) of object-space.  
Each episode consists of 20 steps and each step is a typical tuple $<\mathbf{s}_t, \mathbf{a}_t, \mathbf{r}_t, \mathbf{s}_{t+1}>$ used in reinforcement learning. We use simple but accurate policies to collect data. Accuracy of these data collection policies is above $80\%$.
Figure \ref{fig:obj_spcae_reward_dist}(b) shows how reward is distributed across object space for pick-shed task. Each dot represents a trajectory, and the color represents the total reward for each trajectory. It can be seen, when the object is placed closer to the robot, the reward is high as task is completed in few steps, while it becomes low as the object moves away.

\subsection{Empirical Results and Analysis}

We performed a total of 72 experiments. We performed sequential learning on two task (doublets). Six doublets are possible using data collected for three tasks. These are button-shed, button-drawer, shed-button, shed-drawer, drawer-shed, and drawer-button. For each doublet sequence, we perform 2 sets of experiments, one with SI regularisation and another without SI regularisation. Each set contains 6 experiments by varying  area and density of object-space. Apart from these 72 experiments, we also trained the agent for single tasks using SAC-CQL for reference baseline performance to evaluate forward transfer. We do behaviour-cloning for the initial 5k steps to learn faster as we have limited compute budget.
We use metrics mentioned in \cite{wolczyk2021continual} for evaluating the performance of a continual learning agent. Each task is trained for $\Delta = 100K$ steps. The total number of tasks in a sequence is $N=2$. Total steps $T = 2 \cdot \Delta$. The $i\text{-th}$ task is train from $t \in [(i-1) \cdot \Delta, i \cdot \Delta]$.

\begin{figure*}
\centering
\begin{subfigure}{.5\textwidth}
  \centering
  \includegraphics[width=0.95\linewidth]{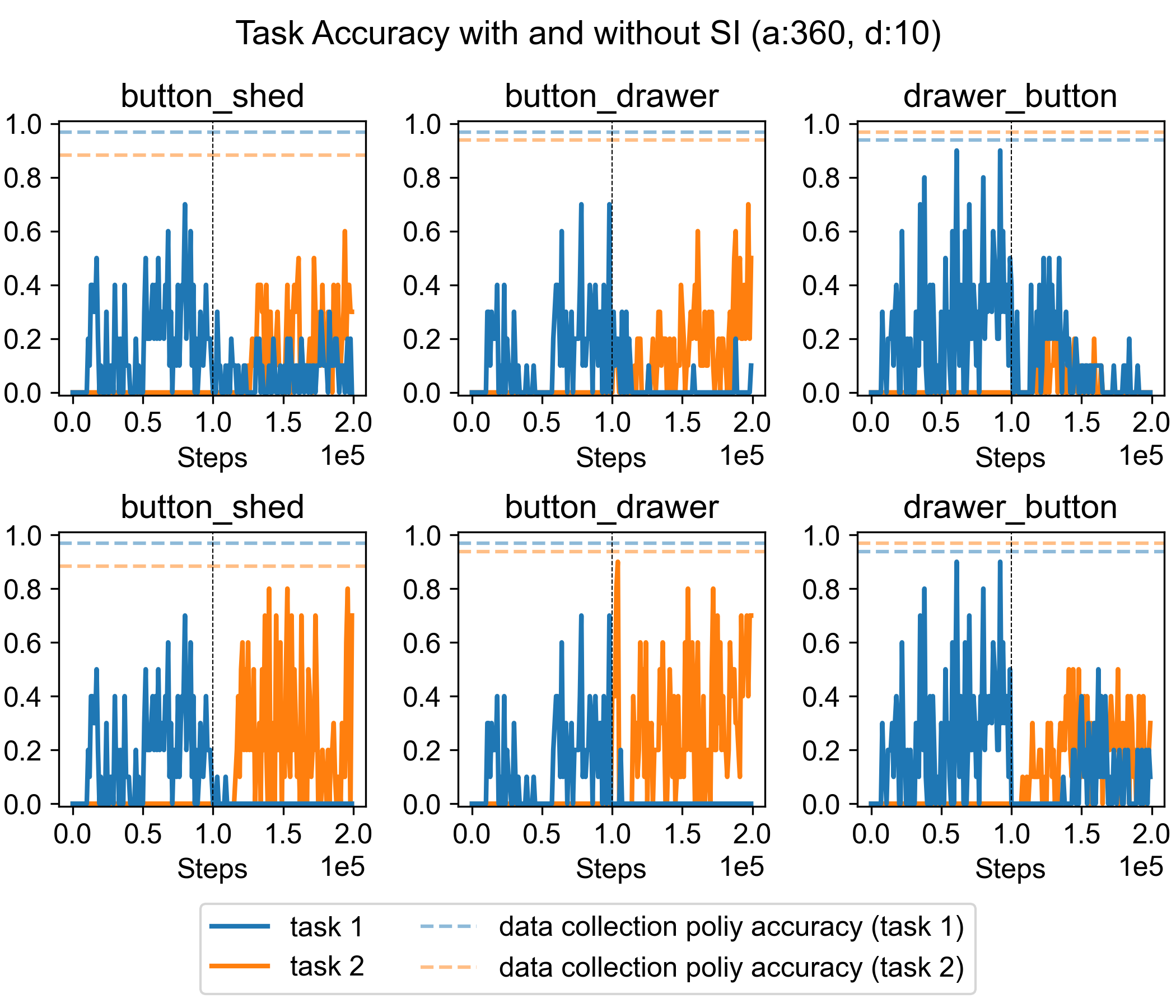}
  \caption{Task Accuracy (area=360, density=10)}
  \label{fig:task_acc_area_360_density_20}
\end{subfigure}%
~
\begin{subfigure}{.5\textwidth}
  \centering
  \includegraphics[width=0.95\linewidth]{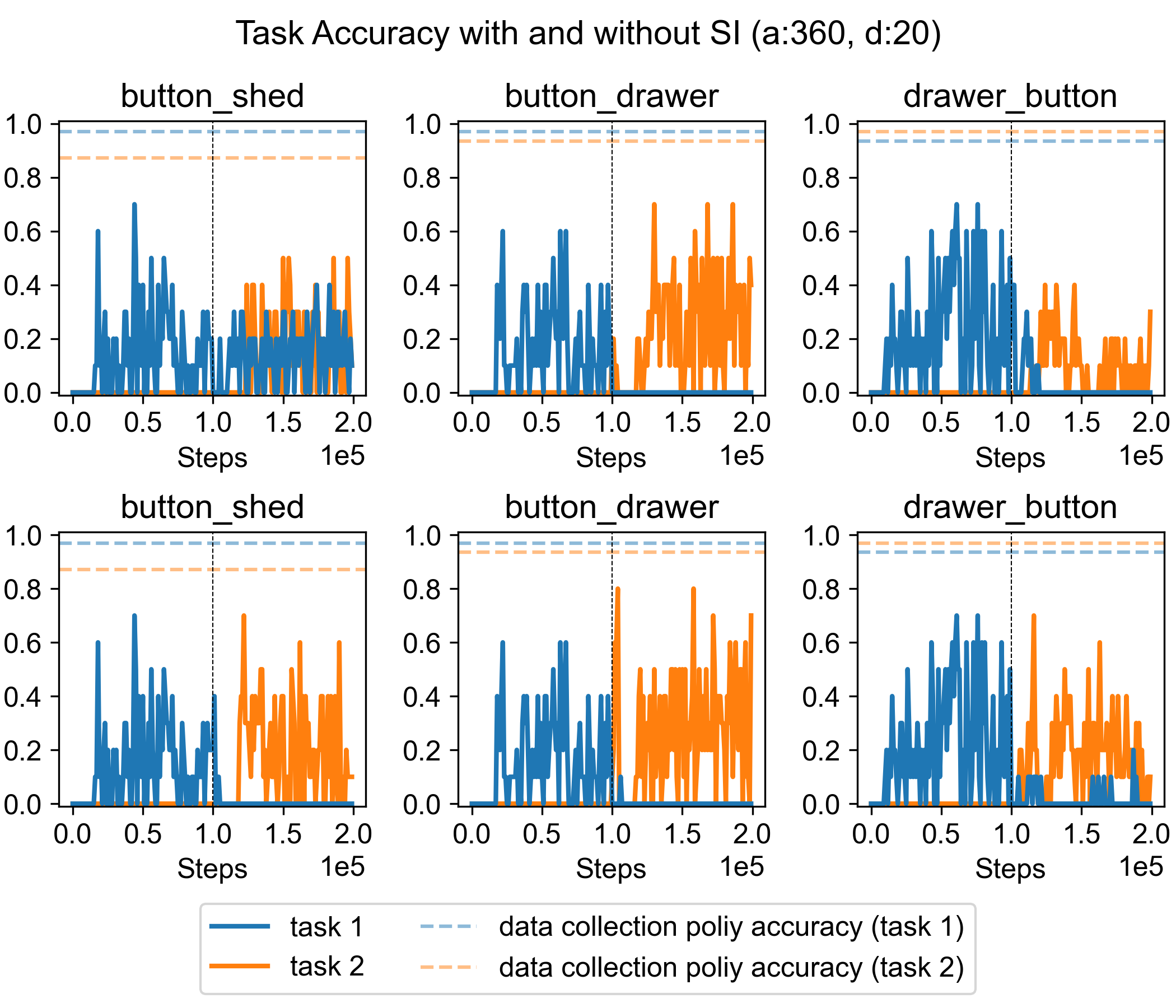}
  \caption{Task Accuracy (area=360, density=20)}
  \label{fig:task_acc_area_360_density_20}
\end{subfigure}
\caption{Task accuracy for tasks button-shed, button-drawer and drawer-button (area=360, density=10,20). Top row is with SI, bottom row is without SI}
\label{fig:task_acc_area_360}
\end{figure*}

\textbf{Task Accuracy:} We evaluate the agent after every 1000 training steps by sampling 10 trajectories from the environment for each task. The accuracy of the agent for a task is defined as the number of successful trajectories out of those 10 trails. Figure \ref{fig:task_acc_area_360} shows the accuracy of three task-sequences (button-shed, button-drawer, drawer-button) over the complete training period of 200k steps for the area size of $40cm^2$ with density of 10 and 20 object configurations per $cm^2$. Top row represents sequential learning with SI while bottom row represents sequential learning without SI. SI is found to be working better as evident by overlapping Task-1 and Task-2 accuracy.
We observed that SI was most helpful in button-shed task doublet due to overlapping nature of these tasks as both these tasks require reaching the object. This shows benefit of using SI for overlapping tasks

\begin{figure*}[!h]
  \centering
  \includegraphics[width=0.95\linewidth]{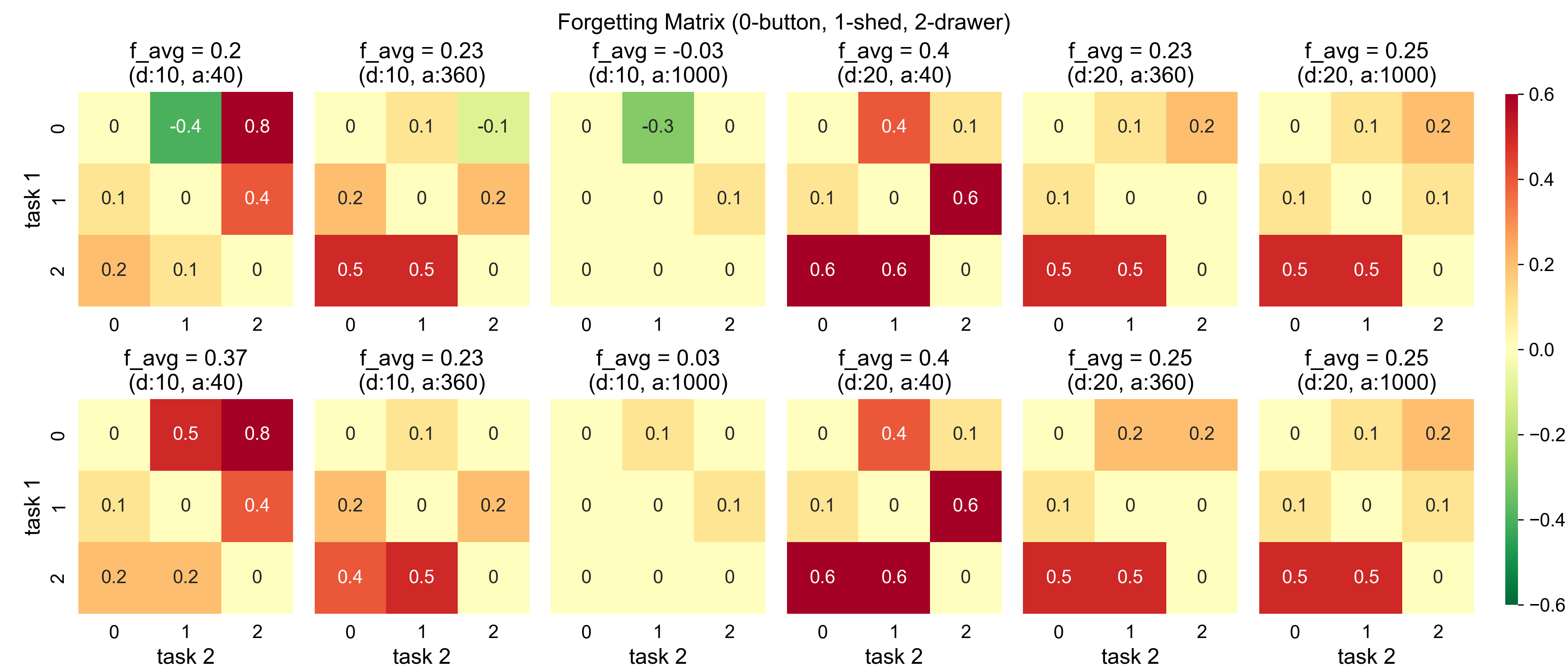}
  \caption{Forgetting Matrix. Top row is with SI regularisation, bottom row is without regularisation}
  \label{fig:f_mat}
\end{figure*}

\textbf{Forgetting:} It measures decrease in accuracy of the task as we train more tasks and defined as $F_i := p_i(i.\Delta)-p_i(T)$. Here, $p_i(t) \in [0,1]$ is success rate of task $i$ at time $t$. Figure \ref{fig:f_mat} shows the forgetting of Task-1 after training Task-2. We can see that SI performed better or equal in all cases. In fact, in some cases, like button-shed forgetting is negative, which means the performance of Task-1 improved after training on Task-2. This indicates knowledge transfer from Task-1 to Task-2. This phenomenon is not seen in case of sequential learning without SI. This clearly indicates that SI helps in reducing catastrophic forgetting. No significant trends are observed in variation of object-space area but forgetting increased with the increase in object-space density. This might be due to the limited compute budget (100K) per task as tasks with more area size and density would require more training to show good results.

\begin{figure*}[h!]
  \centering
  \includegraphics[width=0.95\linewidth]{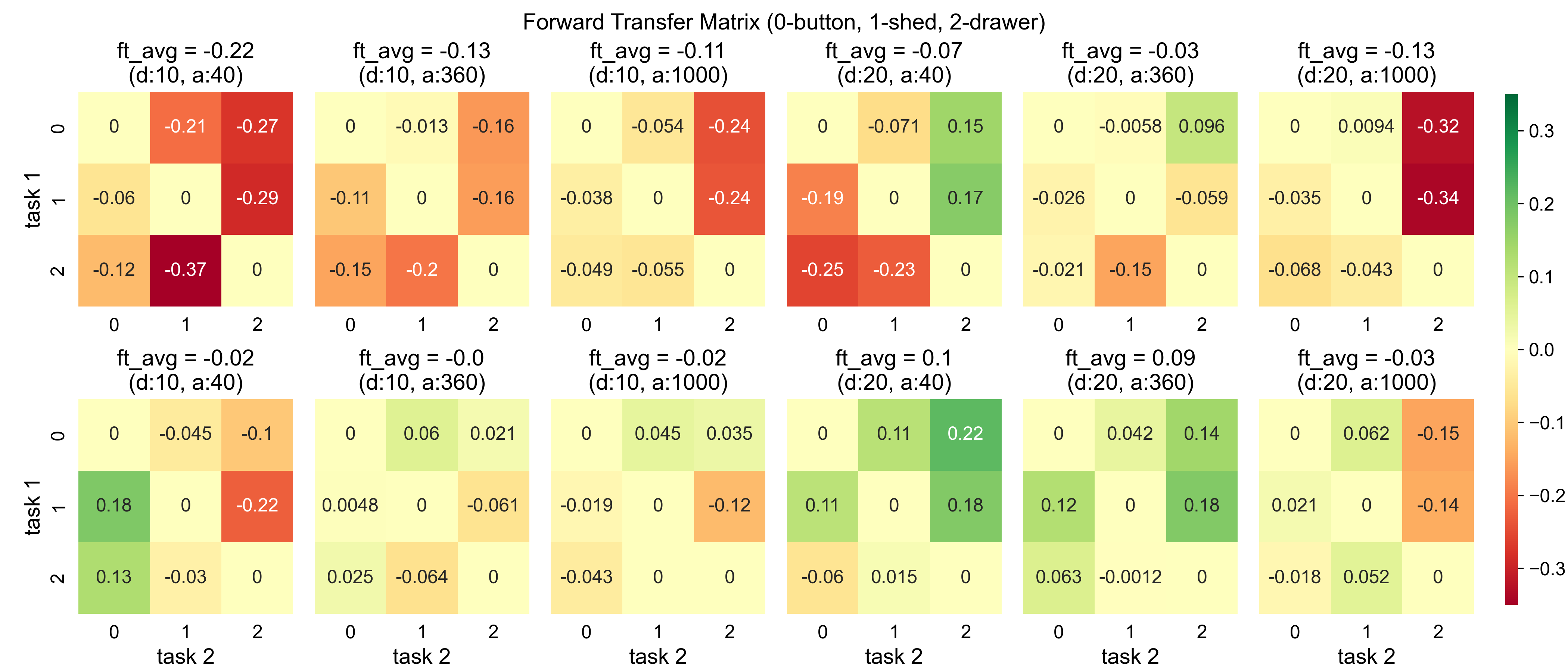}
  \caption{Forward Transfer Matrix}
  \label{fig:ft_mat}
\end{figure*}

\textbf{Forward Transfer:} It measures knowledge transfer by comparing the performance of a given task when trained individually versus learning the task after the network is already trained on previous tasks and defined as 
\begin{equation} \label{eq:forward_transfer}
FT_i := \frac{\text{AUC}_i - \text{AUC}_i^b}{1- \text{AUC}_i^b}, 
\end{equation}
where $\text{AUC}_i = \frac{1}{\Delta} \int_{(i-1) \cdot \Delta}^{i \cdot \Delta} p_i(t)\text{d}t$ represents area under the accuracy curve of task $i$ and $\text{AUC}_i^b = \frac{1}{\Delta} \int_{0}^{\Delta} p_i^b(t)\text{d}t$, represents area under curve of the reference baseline task. $p_i^b(t)$ represent reference baseline performance. Figure \ref{fig:ft_mat} shows forward transfer for Task-2 after it is trained on Task-1. We use single-task training performance as the reference for Task-2 while evaluating forward transfer. We observed that in most cases, training without SI gives a better transfer ratio than training with SI. This may be because of two reasons. Firstly, due to the high value of SI regularisation strength (which is set to 1 for all cases), this restricts movement of weights from the solution of the previous task. This can also be noticed in the form of reduced accuracy levels of Task-2 in the Figure \ref{fig:task_acc_area_360}. The accuracy level of Task-2 are lower as compared to its non-SI counterpart. Although, high regularisation strength helps in reducing catastrophic forgetting but also hinders the ability to learn new-task thus reducing forward-transfer. This highlights the problem of stability-plasticity, any method which tries to make learning more stable to reduce forgetting inadvertently also restricts the flexibility of the connectionist model to learn a new task.

\textbf{Training Time:} Apart from these metrics, we observed that, agent requires on an average 14k, 10k, and 16k steps to achieve its first success on Task-2 when trained directly, sequentially without SI, and sequentially with SI, respectively. This means that the agent learns the task faster when trained sequentially without adding SI regularisation but a little slower when trained sequentially with SI regularisation than directly training the task. This shows another benefit of sequential learning over single task-learning.

Another interesting observation we made in the case of shed-button (area 360, density 20) task. While training for Task-1 (pick shed) agent showed some success on Task-2 (press button) even before getting any success on Task-1 itself. This might be due to the nature of the tasks, as the trajectory of the press button task is common for other task. Therefore, agent has tendency to acquire knowledge for similar tasks. This may also be the result of behaviour-cloning for the initial 5k steps, where the agent tries to mimic the data collection policy for a few initial training steps. Also, we observed that increasing the object space area helps in knowledge transfer, which can be seen by the increase in average forward transfer with area size.

\section{CONCLUSION AND FUTURE WORK}

We investigated catastrophic forgetting and forward knowledge transfer for sequentially learning image-based robotic manipulation tasks by combining a continual learning approach with offline RL framework. We use SAC-CQL as an offline deep RL algorithm with synaptic intelligence (SI) to mitigate catastrophic forgetting. Multi-headed CNN was used to provide knowledge of the current Task-index to the neural-network. We performed a series of experiments with different task combinations and with a varying number of object configurations and densities. We found that SI is useful for reducing forgetting but showed a limited forward transfer of knowledge.

We also found that the ordering of tasks significantly affects the performance of sequential task learning. Therefore, tasks may be chosen in a manner so that the previous task helps in learning the next task as the complexity of tasks increases. This calls for exploring curriculum learning for sequential tasks. Experiments also suggests the importance of prior knowledge for continual learning. Agent trained only with state-action pairs of large number of diverse tasks (even without reward), may provide a better prior knowledge.
Future work will also focus on training tasks with more number of steps to explore more interesting patterns.

\newpage
\printbibliography

\end{document}